\documentclass[journal]{IEEEtran}
\usepackage[utf8]{inputenc}
\usepackage[T1]{fontenc}

\newif\ifisrevision
 \isrevisionfalse

 \newif\ifisrevisionnotfinished
  \isrevisionnotfinishedfalse

\usepackage{times}
\usepackage{epsfig}
\usepackage{graphicx}
\usepackage{amsmath}
\usepackage{amssymb}

\usepackage{multirow,array}
\usepackage{moresize} 
\usepackage{tabularx}
\usepackage{booktabs}  
\usepackage{arydshln}

\usepackage{url}

\usepackage{import}

\usepackage[normalem]{ulem}
\usepackage{url}

\usepackage[usenames,dvipsnames,svgnames,table]{xcolor}
\usepackage{soul}
\usepackage{xspace}
\newcommand{\Note}[1]{}
\renewcommand{\Note}[1]{\hl{[#1]}}  

\ifisrevisionnotfinished
  \ifisrevision
  \newcommand{\revision}[2]{\sout{#1} \textcolor{blue}{#2}} 
  \else
  \newcommand{\revision}[2]{\textcolor{blue}{#2}}
  \fi
\else
\newcommand{\revision}[2]{#2}
\fi

\newcommand{\pytorch}[0]{PyTorch\xspace}
\newcommand{\mtl}[0]{multi-task\xspace}
\newcommand{\Mtl}[0]{Multi-task\xspace}
\newcommand{\dfc}{DFC2018\xspace}
\newcommand{\sener}{MTL-MGDA\xspace}

\newcommand{\lone}{$\mathcal{L}_{1}$\xspace}
\newcommand{\lce}{$\mathcal{L}_{\text{CE}}$\xspace}

\usepackage{xspace}

\newcommand*{\ie}{i.e.\@\xspace}
\newcommand*{\etal}{et al.\@\xspace}
\newcommand*{\soa}{state-of-art\@\xspace}
\newcommand*\rot{\rotatebox{90}}

\usepackage{tikz}

\begin{document}

\title{Multi-Task Learning of\\ Height and Semantics from Aerial Images}

\author{Marcela Carvalho, Bertrand Le Saux, Pauline Trouvé-Peloux, Frédéric Champagnat and Andrés Almansa
\thanks{M. Carvalho, B. Le Saux, P. Trouvé-Peloux and F. Champagnat are with DTIS, ONERA, Universit\'{e} Paris-Saclay, FR-91123 Palaiseau, France}
\thanks{A. Almansa is with MAP5, Universit\'{e} Paris Descartes, FR-75006 Paris, France} }


\maketitle

\begin{abstract}
  Aerial or satellite imagery is a great source for land surface analysis, which might yield land use maps or elevation models. In this investigation, we present a neural network framework for learning semantics and local height together. We show how this joint multi-task learning benefits to each task on the large dataset of the 2018 Data Fusion Contest. Moreover, our framework also yields an uncertainty map which allows assessing the prediction of the model. \revision{}{Code is available at \url{https://github.com/marcelampc/mtl_aerial_images} }.
\end{abstract}

\begin{IEEEkeywords}
Multi-task learning, Aerial imagery, Semantic Segmentation, Single view depth estimation, Neural networks, Deep learning
\end{IEEEkeywords}

\IEEEpeerreviewmaketitle

\section{Introduction \label{sec:intro}}

Aerial imagery has never been so common, even at Very High Resolution (VHR), now that everyone can access images from around the world in any computer. Its automatic analysis is also in progress and has been boosted in the last decade by the tremendous progresses of convolutional neural network (CNNs). It includes spectral analysis, change detection, and two applications which are of particular interest in this study: semantic mapping of the land surface and local height estimation.

Adding semantics to images by creating high-quality land-cover maps is crucial for environment analysis or urban modelling. A standard way to formulate this problem is classification of each pixel, now reframed as semantic segmentation~\cite{paisitkriangkrai_effective_2015, audebert-18isprsj-beyond-RGB}.
Besides, providing the local height in the form of Digital Surface Models (DSMs) is useful for urban planning, telecommunications, aviation, and intelligent transport systems.
 It has been traditionally done by multi-view stereo~\cite{facciolo-MVS-cvprw2015} until that recently, deep learning approaches also offer competitive performances~\cite{lesaux-grsm19-dfc2019announce}. Eventually, \cite{srivastava2017joint} made one step further by combining both tasks through \mtl learning. In part~\ref{sec:soa}, we explore more related work to give insight into these approaches.

We also tackle this problem by using a \mtl deep network that simultaneously estimates both height and semantic maps from a single aerial image \textit{and we show that both tasks benefit from each other}. To reach this goal, our approach builds on powerful models for depth prediction from a single image~\cite{Carvalho2018icip}. Moreover, \revision{it comes with a measure of the model uncertainty~\cite{kendall2015bayesian} which helps understanding success and failure cases.}{we investigate the model uncertainty~\cite{kendall2015bayesian} to bring a new regard to aerial imagery processing and better understand success and failure cases. We also perform a study with most recent \mtl techniques for scene-understanding to analyse their contributions when confronted to the \dfc dataset.}

Notably, we obtain state-of-the-art results using Very-High-Resolution (VHR) imagery only on reference datasets: IEEE GRSS DFC2018~\cite{iadf-19jstars-dfc18} and ISPRS Semantic Labeling~\cite{cramer2010dgpf}. On challenging DFC2018 data, comparing true deep learning methods only, our approach achieves 8\% more accuracy than the winning solution Fusion-Net without post-processing~\cite{yonghao-xu-DFC18winner-igarss18}. 

After the related work in part~\ref{sec:soa}, we will describe our multi-task approach in part~\ref{sec:method} and present results in part~\ref{sec:exp}.

\begin{figure}
  \centering
  \includegraphics[width=.9\linewidth]{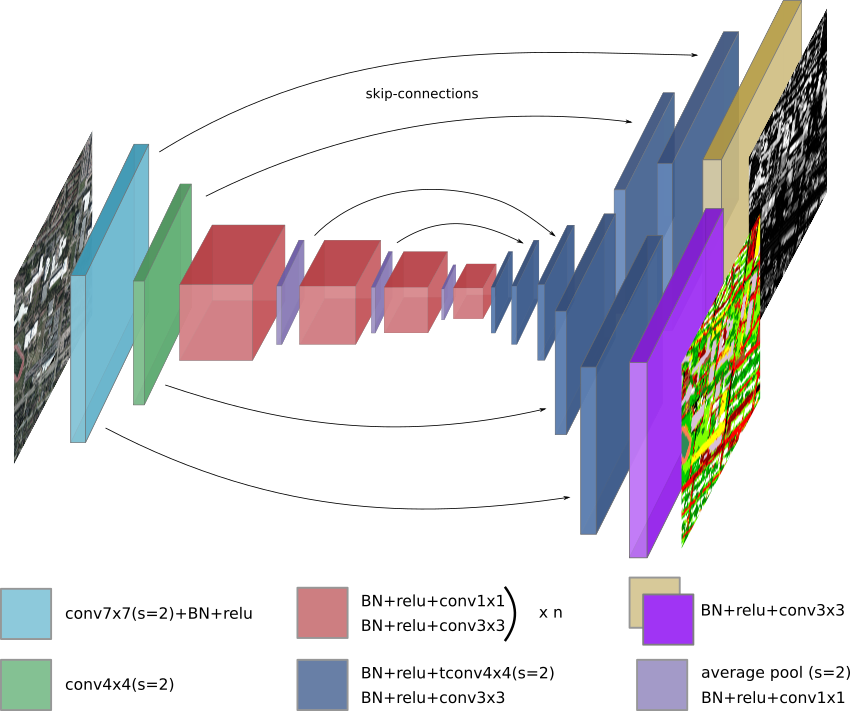}
  \caption{\label{fig:architecture} Architecture of our \mtl model for aerial imagery, based on D3-Net~\cite{Carvalho2018icip}. Left most layers share parameters between all tasks and right most layers are task-specific. Last layer of each decoder differs only on the output number of channels, followed by task evaluation metric (cross-entropy, or \lone, for semantic and height estimation, respectively). }
\vspace{-2mm}
\end{figure}

\section{Related work \label{sec:soa}}


\paragraph{Semantic segmentation} this task consists in giving a class label to each pixel in the image~\cite{brostow-CamVid-PRL2009} and has been commonly carried out in the recent years by Fully-Convolutional Networks (FCNs) since~\cite{long_fully_2015}. In remote sensing, it corresponds to the old problem of land-surface classification~\cite{benediktsson-classif-remote-sensing-TGRS1990} and has been popularized again by recent benchmarks on urban land-use mapping~\cite{cramer2010dgpf, iadf-19jstars-dfc18}. Current state-of-the-art approaches based on FCNs include~\cite{paisitkriangkrai_effective_2015, audebert-18isprsj-beyond-RGB} or~\cite{marmanis-class-edge-ISPRS2016} which combines segmentation with boundary detection. When multi-source data is available, as in the 2018 DFC, 
dedicated network architectures such as Fusion-CNN~\cite{iadf-19jstars-dfc18} can be designed to use this information.

\paragraph{Elevation / depth estimation} the problem at hand here is to estimate the distance between the sensor and the observed scene, which means depth in computer vision or elevation (up to an affine transformation) in remote sensing. For nearly 40 years, stereo or multi-view stereo have been the means of choice, and still give excellent results~\cite{facciolo-MVS-cvprw2015}. However, recently 3D estimation from a single image became popular thanks to the conjunction of availability of image-and-depth datasets with finally powerful-enough neural network models~\cite{Eigen2015, laina2016deeper}. In remote sensing, several networks for predicting elevation were also proposed, first~\cite{srivastava2017joint} then~\cite{ mou2018im2height, ghamisi2018img2dsm, amirkolaee2019height}. In particular, \cite{amirkolaee2019height} uses a ResNet-based FCN to produce the DSM while~\cite{ghamisi2018img2dsm} adds an adversarial loss to improve the likelihood of the synthesized DSM. Some works show how the estimated DSM is an useful additional information for building detection~\cite{mou2018im2height} or semantic segmentation~\cite{ghamisi2018img2dsm}. It is worth noting that the 2019 Data Fusion Contest~\cite{lesaux-grsm19-dfc2019announce} comprises one challenge about \textit{Single-view Semantic 3D Challenge} which should yield to new methods to tackle this problem in remote sensing. With respect to previous methods, our approach jointly learns height and semantics, using only VHR imagery.

\paragraph{\Mtl Learning} It aims at discovering the latent relatedness among tasks to improve generalization. In practice, it leverages the domain-specific information contained in the training signals of related tasks to build a better model which benefits all tasks~\cite{Caruana-MTL-MachineLearning1997}.
 Recent works include the simultaneous prediction of depth, normals and semantic labels~\cite{Eigen2015} or normalized DSM and semantic labels~\cite{srivastava2017joint}. In the latter, the network consists mostly in shared hidden convolutional layers followed by task-specific heads: one fully-connected layer and the appropriate loss. With respect to theirs, our multi-task architecture favors a middle split for the division in two task-specific branches, \revision{which we think is more suitable for tasks as diverse as semantic mapping and DSM regression.}{a more suitable strategy for tasks as diverse as semantic mapping and DSM regression as this gives the model more  specialized layers for each objective.} As Caruana pointed out, backpropagation is one of the mechanisms to discover task relatedness~\cite{Caruana-MTL-MachineLearning1997}. Several recent works have dealt with the balance of the influence of each task during backpropagation. They weight task specific losses according to their intrinsic uncertainty~\cite{kendall-MTL-uncertainty-CVPR2018}, directly the gradient magnitude for GradNorm~\cite{chen-badrinarayanan-GradNorm-ICML2018}, or to Pareto improvements between the conflicting tasks for \sener~\cite{sener-koltun-MTL-as-multi-obj-optim-NeurIPS2018}.
 
\section{Method \label{sec:method}}

\textbf{Network Architecture.}
We adapted D3-Net, an encoder-decoder deep network originally created for depth estimation, to a multi-task architecture by adding a semantic classification decoder. This architecture favors hard parameter sharing: as illustrated in Fig.~\ref{fig:architecture}, the contractive and the early decoder layers are common for both semantics and height estimation. Last layers of the decoders are specific for each objective and generate respectively as many channels as classes for semantics and one channel for height. We also implemented various mechanisms for balancing tasks during optimization following literature in~\cite{chen-badrinarayanan-GradNorm-ICML2018,sener-koltun-MTL-as-multi-obj-optim-NeurIPS2018}

\begin{figure}[t]
  \centering
  \includegraphics[width=0.7\linewidth]{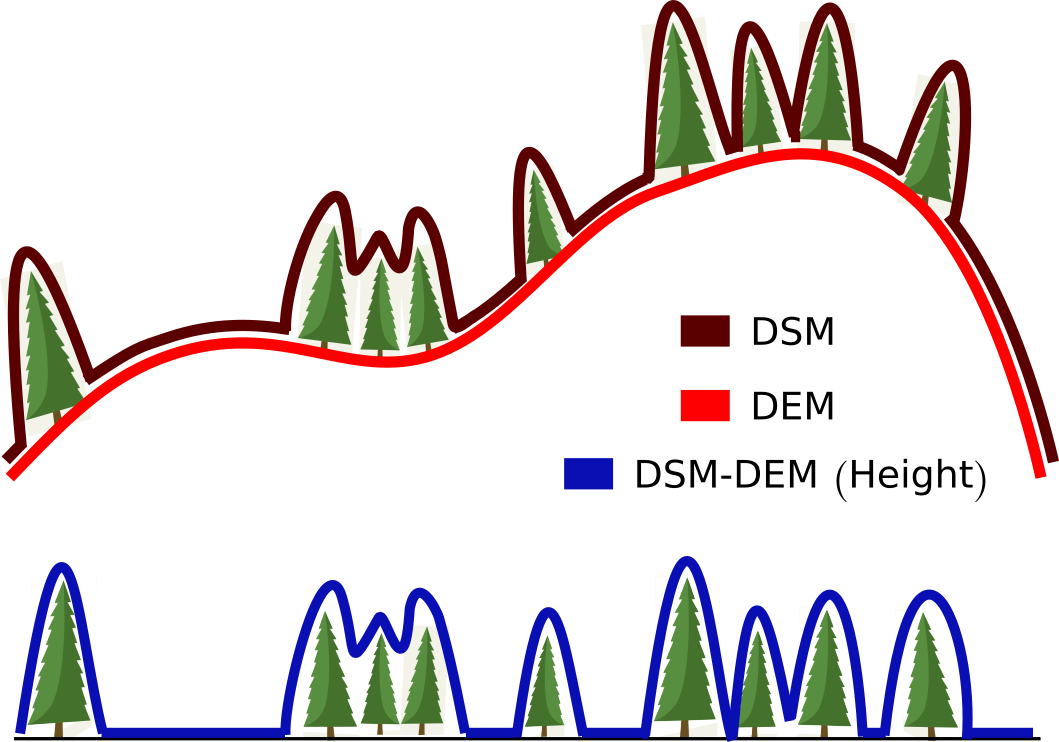}
  \caption{\label{fig:dem_dsm}Height map generation using the Digital Surface Model (DSM) and the Digital Elevation Model (DEM): height as DSM-DEM.}
\vspace{-2mm}
\end{figure}

\textbf{Multi-Objective Loss.}
As discussed in section~\ref{sec:soa}, learning multiple tasks requires to correctly balance each objective's contribution at every training iteration. Indeed, each output is evaluated with a corresponding loss function: we adopt the absolute error (\lone) for height regression and the cross entropy loss (\lce) for semantics evaluation. Thus, when errors are backpropagated, the resulting gradient in the common layers will correspond to the sum of all task gradients. A simple way to control the contribution of each task  consists in multiplying each loss term in the final loss~\ref{eq:mtl_loss} by a scalar $k_{t}$. However, finding the optimal values for each $k_{t}$ is still challenging. 
\begin{equation}
 \label{eq:mtl_loss}
  L_{final}=\sum_{t=1}^{T}k_{t}\mathcal{L}_{t}
\end{equation}

Consequently, many methods have been proposed to effectively estimate $k_{t}$ in order to converge common parameters values to the best model for all tasks. Here, we propose to evaluate these methods in the context of aerial imagery. In the following, we explain the main idea of each approach. 

\begin{figure*}[th]
  \centering
  \includegraphics[width=1.0\linewidth]{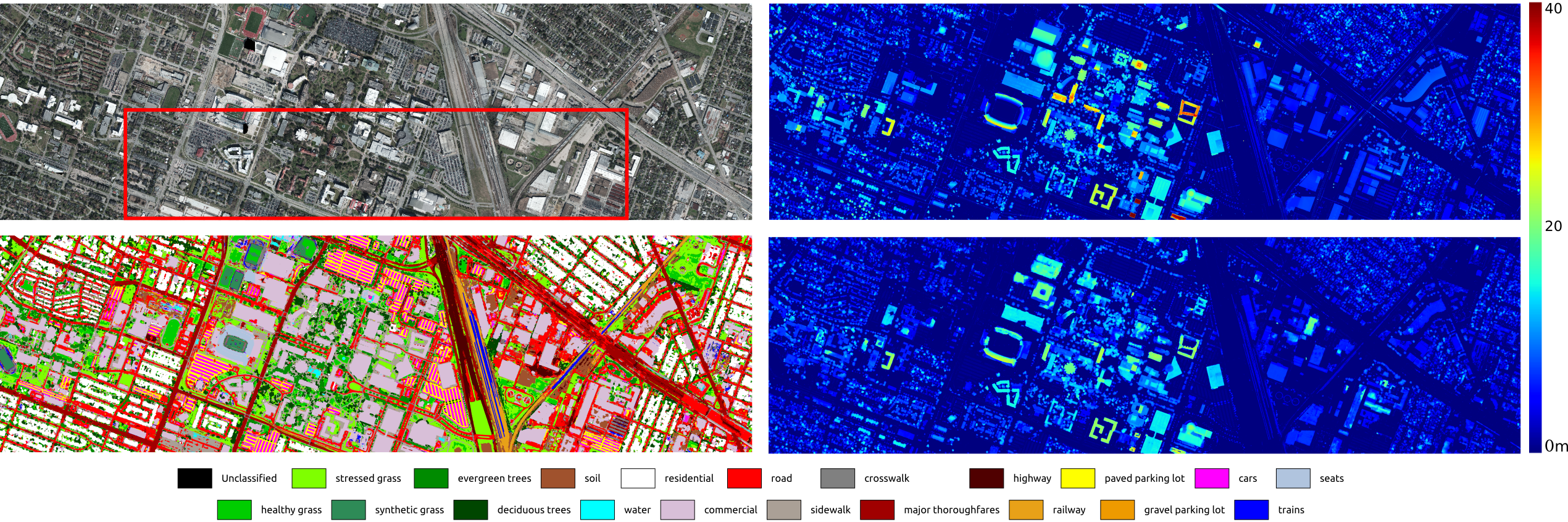}
  \caption{\label{fig:dfc_prediction_overall} Results on the \dfc dataset trained with equal weights (best results). Top row shows RGB image and height ground-truth, bottom row semantic prediction and height estimate. The training area is delimited by a red rectangle on the RGB image.}
\end{figure*}

\paragraph{Equal weights} is the most common approach and consists on weighting all losses uniformly. This approach does not handle cases when the training errors have different scales. In consequence some tasks can be dominant and predictions may be degraded for the other ones. However, this technique can still be effective in the case we can not appropriately measure the best contribution of each task to the global model.

\paragraph{GradNorm~\cite{chen-badrinarayanan-GradNorm-ICML2018}} dynamically learns the scaling factors with respect to the gradients of the last common layer and the rate balance, defined as the relative inverse training rate for each task. By directly modifying gradient magnitudes with learnable parameters, this method does not rely on empirical values for $k_{t}$. 

\paragraph{\sener~\cite{sener-koltun-MTL-as-multi-obj-optim-NeurIPS2018}} Sener~\etal proposed to adapt the Multiple Gradient Descent Algorithm (MGDA) to the multi-objective optimization. This approach uses the Frank-Wolfe algorithm~\cite{jaggi_frankwolfe_icml2013} to find a common descent direction to the gradients of the shared layers at each iteration. Besides that, the paper also proposes to reduce memory use by applying the MGDA to the upper-bound \textit{\sener-UB} of the objective by means of the gradient of task losses with respect to the intermediate representation on the last common layer.

\section{Experiments \label{sec:exp}}

In this section, we first describe the two datasets and the metrics used in this study. Experiments are as follows. First, we compare the general approach of \mtl using equal weights to single-task objectives and also to current \soa results on the proposed datasets to validate our approach and architecture. Then, we extent our study by analyzing the uncertainty map of the proposed model, according to~\cite{kendall2017uncertainties}. Finally, we compare various flavors of recent \soa \mtl in this context.

\vspace{-2mm}
\subsection{Datasets and Metrics}

The \textbf{ISPRS Vaihingen}~\cite{cramer2010dgpf} dataset comprises IRRG (Infra-Red, Red and Green) images at 9cm / pixel, DSM, and 2D and 3D semantic labeled maps for urban classification and 3D reconstruction. It contains 33 patches of different sizes, of which 16 images are used for training and the remaining 17 are used for testing. Semantic maps were annotated with 6 classes including impervious surfaces, building, low vegetation, tree, car and clutter/background. We ignore this last class during training and testing. As in~\cite{srivastava2017joint}, for height estimation we adopt the normalized DSMs (nDSM) from~\cite{gerke2015use-of-stair-vision-isprs}.

The \textbf{2018 Data Fusion Contest} (\dfc~\cite{iadf-19jstars-dfc18}) dataset is a collection of multi-source optical imagery over Houston, Texas. In particular, it contains Very High Resolution (VHR) color images resampled at 5cm / pixel, hyperspectral images, and LiDAR-derived products such as DSMs, and Digital Elevation Models (DEMs) at a resolution of 50cm / pixel. The original \dfc dataset does not include height maps, thus we generate them by substracting the DEM to the DSM, as illustrated in Fig.~\ref{fig:dem_dsm}. A 20-class, handmade ground-truth exists: 4 tiles (corresponding to the VHR images in the red frame in Fig.~\ref{fig:dfc_prediction_overall}) are available for training, while 10 tiles remain undisclosed for evaluation on the the DASE website~\footnote{GRSS Data and Algorithm Standard Evaluation website:~\url{http://dase.grss-ieee.org/}}.

\textbf{Pre-processing.} For both datasets, we perform training using 320x320 crops from the original images. As mentioned, RGB images from \dfc are 10 times bigger than the height and semantic models. So, we perform training with two different strategies: first, to deal with VHR images, we upsample the height and semantic maps to the same resolution of the input image before performing crops; second, to speed up training and testing, we downsample the RGB images by a factor of 10. We refer to these pre-processing strategies as VHR \Mtl and LR (low-resolution) \Mtl. 

\textbf{Data augmentation.} To improve generalization, we perform the following online data augmentation: random crops from original tiles, rotation, horizontal and vertical flips.

\textbf{From crops to tiles.} Inference is implemented using a Gaussian prior over patches to avoid a checkerboard effect on the output. We predict patches sequentially with a stride smaller than the window size and weight overlapping areas with a 2D Gaussian map. Results are improved when using bigger windows and small strides as we can leverage more information from neighbor patches. For our experiments, we use a test window of 1024 and a stride of 256. When generating VHR outputs, these are posteriorly downsampled to compare to ground truth maps.

Training is performed with \pytorch~\cite{paszke2017pytorch} framework, we used Adam~\cite{kingma2014adam} as our optimizer with learning rate of 2e-4 and we train our model with a Nvidia GTX 1080 GPU.

\textbf{Metrics.} To evaluate our models, we use common metrics from~\cite{Eigen2015} and ~\cite{srivastava2017joint}. For height estimation, we use the mean absolute error (mae): {\scriptsize $\frac{1}{N}\sum_{i=0}^{N}|d_{i}-\hat{d}_{i}|$}; the mean squared error (mse), {\scriptsize $\frac{1}{N}\sum_{i=0}^{N}{\left\Vert d_{i}-\hat{d}_{i}\right\Vert}^{2}$}; and the root mean squared error (rmse), {\scriptsize $\sqrt{\frac{1}{N}\sum_{i=0}^{N}{(d_{i}-\hat{d}_{i})}^2}$}. For classification, we use overall accuracy (OA), average accuracy (AA) and Kappa, as in~\cite{iadf-19jstars-dfc18}.

\vspace{-2mm}
\subsection{Improving Height and Semantic Estimation with \Mtl}
\revision{By using \textit{equal weights}, we assume that tasks have close error scales. This choice is faster and less memory consuming than \soa techniques (\ie, GradNorm, \sener).}
{In this experiment, we use the original D3-Net with the corresponding task decoder for single task training, and the proposed model with \textit{equal weights} for \mtl.}

\begin{figure*}[th]
  \centering  
  \begin{tabular}{c@{\hskip 4pt}c@{\hskip 4pt}c@{\hskip 4pt}c@{\hskip 4pt}c}
   \includegraphics[width=0.19\textwidth]{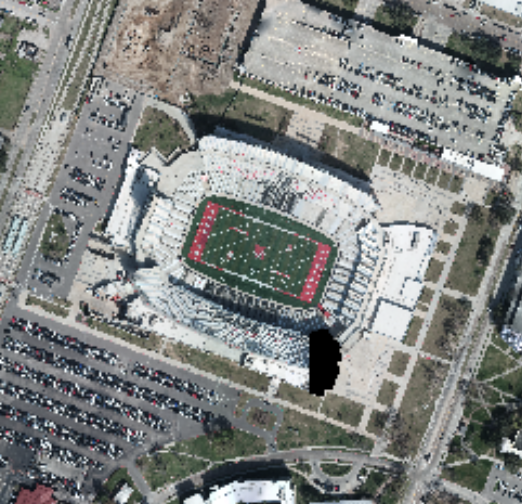} & \includegraphics[width=0.19\textwidth]{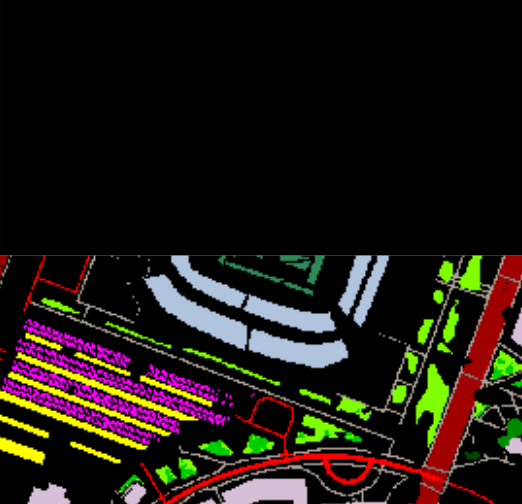} & 
   \includegraphics[width=0.19\textwidth]{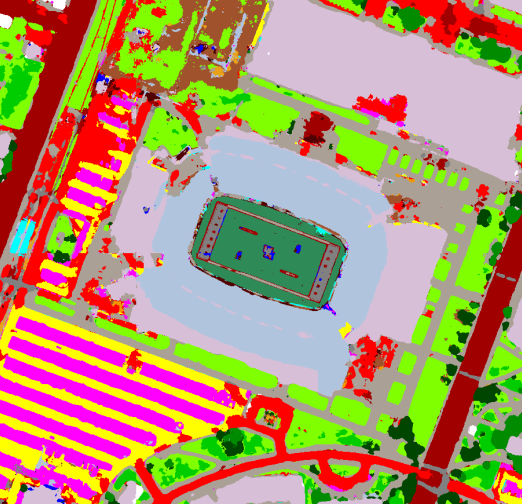} & \includegraphics[width=0.19\textwidth]{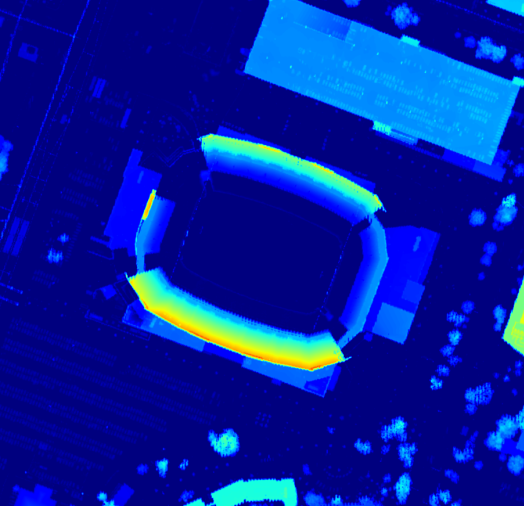} &
   \includegraphics[width=0.19\textwidth]{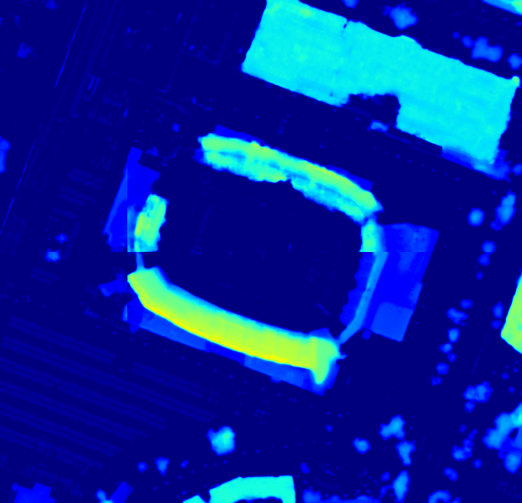} \\
   \includegraphics[width=0.19\textwidth]{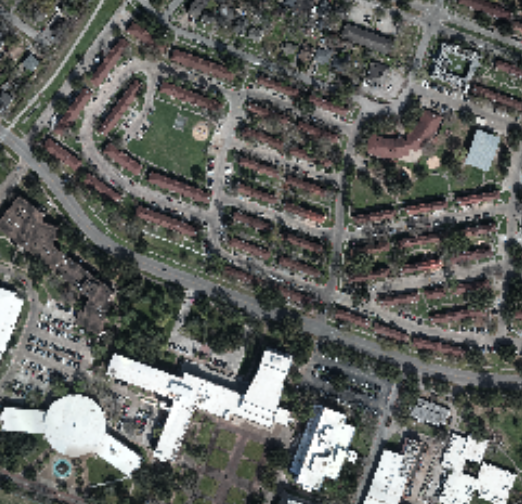} & \includegraphics[width=0.19\textwidth]{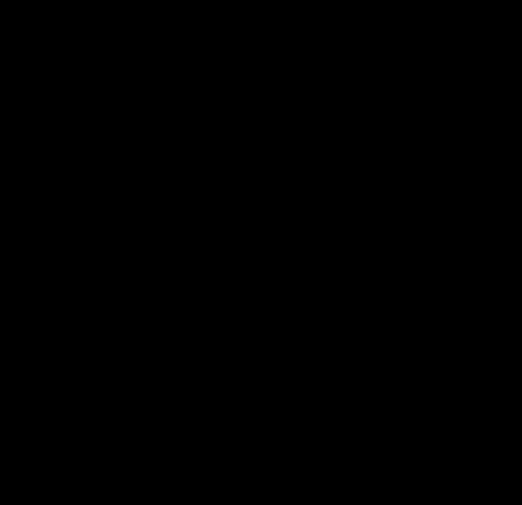} & 
   \includegraphics[width=0.19\textwidth]{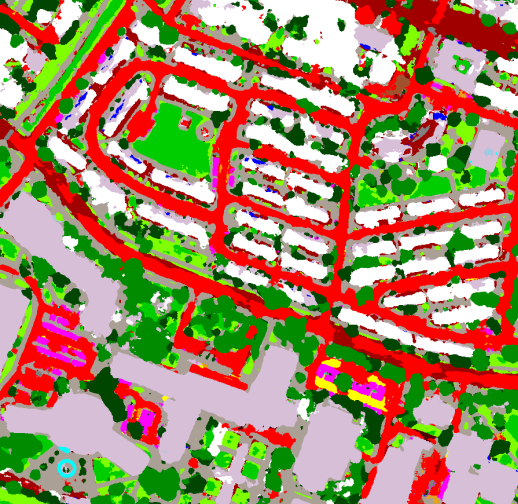} & \includegraphics[width=0.19\textwidth]{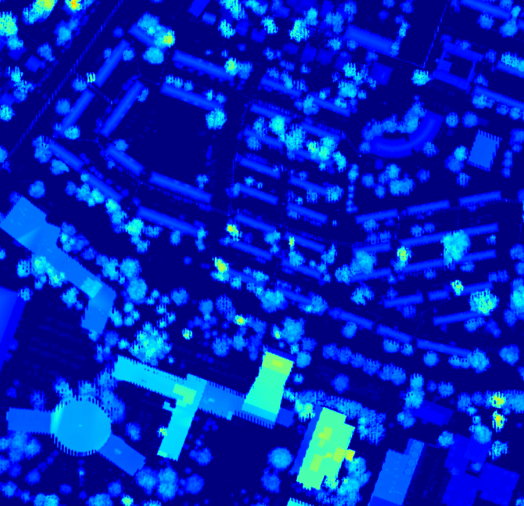} &
   \includegraphics[width=0.19\textwidth]{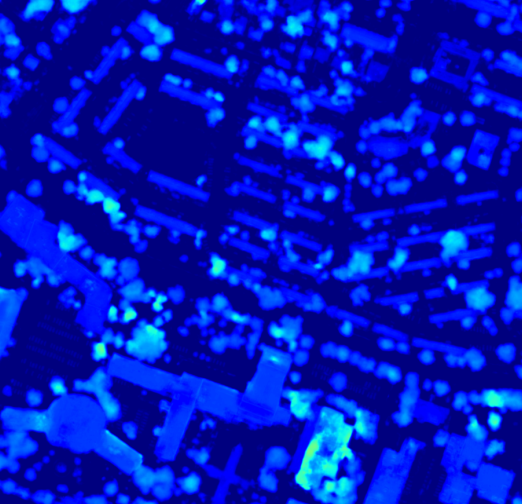} \\
\end{tabular}
\centering
  \caption{\label{fig:dfc_prediction_zoom} Crop areas over the DFC2018 dataset. From left to right, input RGB image, semantic ground-truth and prediction (black is no information), height ground-truth and prediction. Top row show the Houston University stadium and bottom row shows a residential area.}
\end{figure*}

\begin{table}[t]\centering
  \caption{\label{table:results_dfc} Comparison of \dfc height and semantics predictions with state-of-art approaches, single and multi-task models.}
  \footnotesize
  \scalebox{0.9}{
  \begin{tabular}{c@{\hspace{2mm}}c@{\hspace{1mm}}cc c@{}c@{}c@{\hspace{1mm}}c}\toprule
    {\multirow{2}{*}{}} & \multicolumn{3}{c}{Height Errors} & \multicolumn{3}{c}{Semantic Errors} & {\multirow{2}{*}{Time$^{1}$(s)}} \\
   & mae$\downarrow$ & mse$\downarrow$ & rms$\downarrow$ & OA(\%)$\uparrow$ & AA$\uparrow$ & Kappa$\uparrow$ &  \\  \midrule
   Cerra*~\cite{cerra-dfc2018-IGARSS2018}  & - & - & - & 58.60 & 55.60 & 0.56 & - \\
   Fusion-FCN*~\cite{yonghao-xu-DFC18winner-igarss18} & - & - & - & 63.28 & - & 0.61 & - \\
   Fusion-FCN~\cite{yonghao-xu-DFC18winner-igarss18} & - & - & - & 80.78 & - & 0.80 & - \\
  \multicolumn{8}{l}{*learning only, without post-processing}\\
   \midrule
  VHR Single task  & 1.480 & 9.544 & 3.000 & 73.40 & 67.82 & 0.72 & - \\
  VHR \Mtl & 1.263 & 7.279 & 2.599 & 74.44 & 68.30 & 0.73 & 7.74$.10^2$\\ 
  \hdashline[1pt/1pt]
  LR \Mtl  & 1.513 & 9.341 & 2.970 & 64.70 & 58.85 & 0.63 & 7.82 \\ 
  \bottomrule
  \multicolumn{8}{l}{$^{1}$Mean test time per image (std. deviation)}\\
  \end{tabular}
  }
\vspace{-4mm}
\end{table}

\begin{table}[h]\centering
  \caption{\label{table:results_vaihingen} Comparison of \Mtl \soa and our architecture adopting an equal weight approach.}
  \footnotesize
  \scalebox{1.0}{
    \begin{tabular}{c@{\hspace{2mm}}c@{\hspace{2mm}}cc c@{}c@{}c@{\hspace{2mm}}c}\toprule
      {\multirow{2}{*}{}} & \multicolumn{3}{c}{Height Errors} & \multicolumn{3}{c}{Semantic Errors} \\
   & mae$\downarrow$ & mse$\downarrow$ & rms$\downarrow$ & OA*(\%)$\uparrow$ & AA**$\uparrow$ & Kappa$\uparrow$ \\  \midrule
   Srivastava~\cite{srivastava2017joint} & 0.063 & - & 0.098 & 78.8\% & 73.4\% & 71.9\%  \\
   \midrule
   Single task & 0.039 & 0.005 & 0.067 & 87.4\% & 84.4\% & 75.0\% \\ 
  \Mtl & 0.045 & 0.006 & 0.074 & 87.7\% & 85.4\% & 75.9\% \\
  \bottomrule
  \end{tabular}
  }
\vspace{-3mm}
\end{table}

If we focus on the bottom lines of Table~\ref{table:results_dfc}, we observe that performances are improved for both objectives by the multi-task model if compared to the single-task. It has the advantage of learning complementary features, using less parameters compared to single models for each task. 

In the upper lines, 
we observe \soa results for \dfc. Please note these methods use multi-source data from DSM, DEM, Hyperspectral image and VHR RGB as input to estimate semantic maps only. For fairness of comparison, results with a $*$ refer to methods without ad-hoc detector nor post-processing.
It appears our model overcomes past learning-based approaches by 10 percentual points on OA.

The above results are inferred in nearly 13 minutes for each $10^4 \times 10^4 \mathrm{pixels}$ tile, when using the inference proposed in section~\ref{sec:method}. For large batch processing, this time can even be reduced by using the LR model, which reduces time to nearly 10 seconds per tile at the cost of losing efficiency.

We can also observe results in Fig.~\ref{fig:dfc_prediction_overall} and crops for specific regions in Fig.~\ref{fig:dfc_prediction_zoom}. In general, the network produces nearly accurate results for ground, residential buildings and vegetation, while some structures are more challenging, like high buildings or stadiums. These classes have various shapes, colors and heights. Thus, it is hard to estimate precise height values from bird-view. Semantics are detailed, with even plastic seats, playground or concrete elements in the stadium. 

Results with Vaihingen dataset are in Table~\ref{table:results_vaihingen}. We observe that our performances overcome Srivastava~\etal\cite{srivastava2017joint}. This is likely due to a better network with skip-connections and an earlier split between task-specific decoders. It is worth noting that our \mtl approach only improves semantic classification if compared to single-task models. In~\cite{srivastava2017joint}, none of the tasks was improved by \mtl. Possible reason is that Vaihingen does not have much variance between train and test sets, so even single task models can overfit and also be performing during inference. Actually, \mtl improves generalization by leveraging complementary information between tasks. 

\begin{figure*}[t]
  \centering
\vspace{-2mm}
  \centering
  \includegraphics[width=\linewidth]{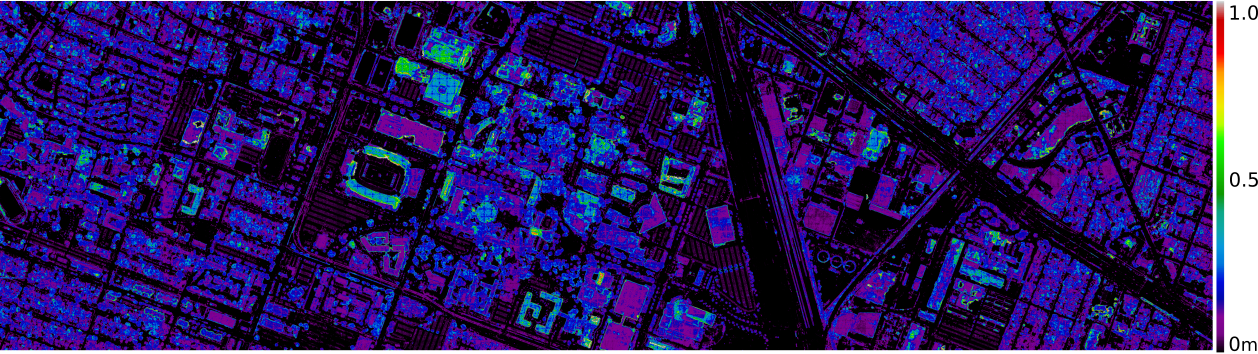}
  \caption{\label{fig:dfc_uncertainty} Uncertainty map of height (standard deviation of model predictions).}
\vspace{-2mm}
\end{figure*}

\subsection{Uncertainty}

In addition to error measures, \cite{kendall2015bayesian} proposed to evaluate the uncertainty of the network, which accounts for the ignorance of the model parameters with respect to the input images. 
To perform this analysis, we follow the original paper and keep dropout layers active during inference. For each tile, we generate 30 samples from which we calculate the standard deviation of the predictions. We perform this test for height estimation only. 
The results in Fig.~\ref{fig:dfc_uncertainty} allows us to understand which zones of the input are the most challenging to the network. Contours present high variance and are challenging. 
So are high buildings in general: indeed a plane rooftop appears the same whatever its altitude. We also note that trees are quite uncertain even if predictions were good: this is a difficult class due to texture variance or deciduousness. 
\vspace{-2mm}
\subsection{Comparison between \Mtl Methods}

In this section, we compare the classic approach with equal weights to \soa methods for \mtl learning. 
These techniques were previously tested on datasets for digit classification, multi-label classification, urban outdoor and indoor scene understanding (Cityscape~\cite{Cordts2016Cityscapes} and NYUv2~\cite{Silberman2012NYUv2}).
We now test them for the first time on VHR aerial images.

As the chosen methods originally rely on architectures without skip connections, we perform experiments with and without these features for best comparison. We observe results for the mentioned methods in Tables~\ref{table:results_exp_soa_mtl_vaihingen} and \ref{table:results_exp_soa_mtl_dfc18}.

\revision{Uniform weighting surprisingly leads to best performances. Indeed, many zones of the training set have sparse information for semantics, leading to the methods based on network gradients to fail when optimizing scales $k_i$. }{From our experiments, we observed that aerial imagery requires less context for the objective tasks than scene-parsing datasets in~\cite{chen-badrinarayanan-GradNorm-ICML2018,sener-koltun-MTL-as-multi-obj-optim-NeurIPS2018}. We believe that subtle \mtl techniques are more prone to better results on these kind of datasets,. Also, in the case of \dfc, semantic annotations are very sparse and gradient values are impacted, which compromising other \mtl methods.} 

\begin{table}[h]\centering
  \caption{\label{table:results_exp_soa_mtl_vaihingen} Comparison of different \mtl approaches from the \soa with Vaihingen dataset}
  \footnotesize
  \scalebox{1.0}{
  \begin{tabular}{@{\hspace{-0.1mm}}c@{\hspace{0.9mm}}c@{\hspace{2mm}}c@{\hspace{2mm}}c@{\hspace{2mm}}c c@{}c@{}c@{}c}\toprule
    & {\multirow{2}{*}{}}  & \multicolumn{3}{c}{Height Errors$\downarrow$} & \multicolumn{3}{c}{Semantic Errors$\uparrow$} \\
  & MTL Method & mae & mse & rms & OA(\%) & AA(\%) & Kappa \\  \midrule
  & \sener-UB~\cite{sener-koltun-MTL-as-multi-obj-optim-NeurIPS2018}  & 0.042 & 0.006 & 0.075 & 81.9\% & 66.0\% & 55.8\%\\ 
  & GradNorm~\cite{chen-badrinarayanan-GradNorm-ICML2018}             & 0.044 & 0.006 & 0.074 & 87.3\% & 84.2\% & 74.3\%\\ 
  \rot{\rlap{no skip}}
  & Equal Weights                                                     & 0.047 & 0.006 & 0.076 & 87.3\% & 84.2\% & 74.6\%\\ 
  \midrule
& \sener~\cite{sener-koltun-MTL-as-multi-obj-optim-NeurIPS2018} & 0.042 & 0.007 & 0.079 & 85.8\% & 81.4\% & 71.2\%\\ 
& GradNorm~\cite{chen-badrinarayanan-GradNorm-ICML2018}         & 0.040 & 0.005 & 0.068 & 87.4\% & 85.1\% & 75.4\%  \\ 
\rot{\rlap{~skip}}
& Equal Weights                                                 & 0.043 & 0.006 & 0.073 & 87.5\% & 84.9\% & 75.5\% \\
  \bottomrule
  \end{tabular}
  }
\end{table}

\begin{table}[h]\centering
  \caption{\label{table:results_exp_soa_mtl_dfc18} Comparison of different \mtl \soa approaches with VHR input images from \dfc dataset.}
  \footnotesize
  \scalebox{1.0}{
    \begin{tabular}{@{\hspace{-0.1mm}}c@{\hspace{0.9mm}}c@{\hspace{2mm}}c@{\hspace{2mm}}c@{\hspace{2mm}}c c@{}c@{}c@{}c}\toprule
      & {\multirow{2}{*}{}}  & \multicolumn{3}{c}{Height Errors$\downarrow$} & \multicolumn{3}{c}{Semantic Errors$\uparrow$} \\
  & MTL Method & mae & mse & rms & OA(\%) & AA(\%) & Kappa \\  \midrule
  & \sener-UB~\cite{sener-koltun-MTL-as-multi-obj-optim-NeurIPS2018}  & 1.475 & 9.911 & 3.047 & 52.98 &	47.59 &	0.50\\ 
  & GradNorm~\cite{chen-badrinarayanan-GradNorm-ICML2018} & 1.394 & 8.886 & 2.857 & 58.00 & 54.23 &	0.56 \\ 
  \rot{\rlap{no skip}}
  & Equal Weights & 1.520 & 8.589 & 2.826 & 58.26 &	54.74 &	0.56 \\ 
  \midrule
& \sener~\cite{sener-koltun-MTL-as-multi-obj-optim-NeurIPS2018} & 1.303 & 7.415 & 2.627 & 59.13 & 55.53 & 0.57 \\ 
& GradNorm~\cite{chen-badrinarayanan-GradNorm-ICML2018} & 1.340 & 7.898 & 2.743 & 63.07 &	58.92 &	0.61 \\ 
\rot{\rlap{~skip}}
& Equal Weights & 1.263 & 7.279 & 2.599 & 74.44 & 68.30 & 0.73 \\
  \bottomrule
  \end{tabular}
  }
\end{table}

\section{Conclusions}
In this work, we have shown that \mtl learning methods work really well on aerial imagery and may lead to better results when compared to single-task techniques. We proved that complementary features from each objective can be learned by a deep model to improve performance independently. Our experiments on \dfc show that a model with less input data and no special post-processing can lead to results comparable to the much complex \soa results. Thus, this framework can be easily adopted for urban modelling without the any complementary information. However, experiments with very recent \mtl variants showed that surprisingly the simple equal-weight approach leads to best performances. Maybe subtle \mtl methods require larger and densely labeled datasets. 
Hence, we will further our work to understand the mechanisms of \mtl. 

\ifCLASSOPTIONcaptionsoff
  \newpage
\fi

\bibliographystyle{IEEEtran}
\bibliography{grsl}

\end{document}